\documentclass[letterpaper]{article} 
\usepackage[preprint]{aaai2027}
\usepackage[hyphens]{url}  
\usepackage{graphicx} 
\usepackage{natbib}  
\usepackage{caption} 
\usepackage{algorithm}
\usepackage{algorithmic}

\usepackage{amssymb}
\usepackage{booktabs}
\usepackage{multirow}
\usepackage{amsmath}
\usepackage{makecell}
\usepackage{array}

\usepackage{printlen}
\uselengthunit{in}

\usepackage{newfloat}
\usepackage{listings}
\DeclareCaptionStyle{ruled}{labelfont=normalfont,labelsep=colon,strut=off} 
\floatstyle{ruled}
\newfloat{listing}{tb}{lst}{}
\floatname{listing}{Listing}

\usepackage{booktabs}
\usepackage{colortbl}
		\title{T-JEPA: A Temporal Joint-Embedding Predictive Architecture for Learning Better Remote Sensing Representations}
		\author{
			Bowen Peng,
			Li Liu\corresponding,
			Yongxiang Liu\corresponding,
			Weijie Li,
			Jie Zhou,
			Zhen Liu
		}
		\affiliations{
			National University of Defense Technology, Changsha, China\\
			\{pbow16, liuli\_nudt, zhoujie\_, zhen\_liu\}@nudt.edu.cn, \{lwj2150508321, lyx\_bible\}@sina.com
		}

\begin{document}
			
			\maketitle
			
			\begin{abstract}
			Earth observation (EO) data provide rich temporal supervision, yet existing remote sensing foundation models mainly exploit sequential observations through imposing predefined pairwise relations or aggregating holistic reconstruction context. We seek to further exploit the sparse and nonuniform temporal sampling inherent in EO sequences as supervisory signals. To this end, we propose T-JEPA, a temporal joint-embedding predictive architecture that learns time-gap-conditioned latent transitions. A shared single-frame encoder processes each observation, while a temporal predictor estimates the complete target latent field from a masked source latent representation and the actual elapsed time. Across multiple temporal intervals, these predictive constraints organize observed states into structured latent trajectories. Asymmetric metadata injection mitigates shortcut learning, and direct supervision across multiple temporal scales proves more effective than recursively rolling out intermediate states. In parallel, masked pixel reconstruction provides complementary supervision for preserving spatial details. Under matched pre-training data and throughput, T-JEPA achieves leading transfer performance on both static and temporal tasks. Analyses further reveal that T-JEPA learns representations with time-gap-dependent transition predictability and coherent latent dynamics, while maintaining strong cross-period consistency, representation diversity, and semantic discriminability.
			\end{abstract}
			
						
			\section{Introduction}
			
			Earth observation (EO) data are sequential records of our evolving planet. Repeated acquisitions reveal regularities in seasonal, ecological, and anthropogenic processes, providing valuable self-supervision signals for remote sensing representation learning from large-scale unlabeled archives. Representations that capture such land-surface dynamics can encode more than the appearance of isolated observations; by characterizing both the persistence and temporal evolution of geographic entities, they are expected to transfer more effectively to downstream tasks.
			
			\begin{figure}[tb]
				\centering
				\includegraphics[width=\linewidth]{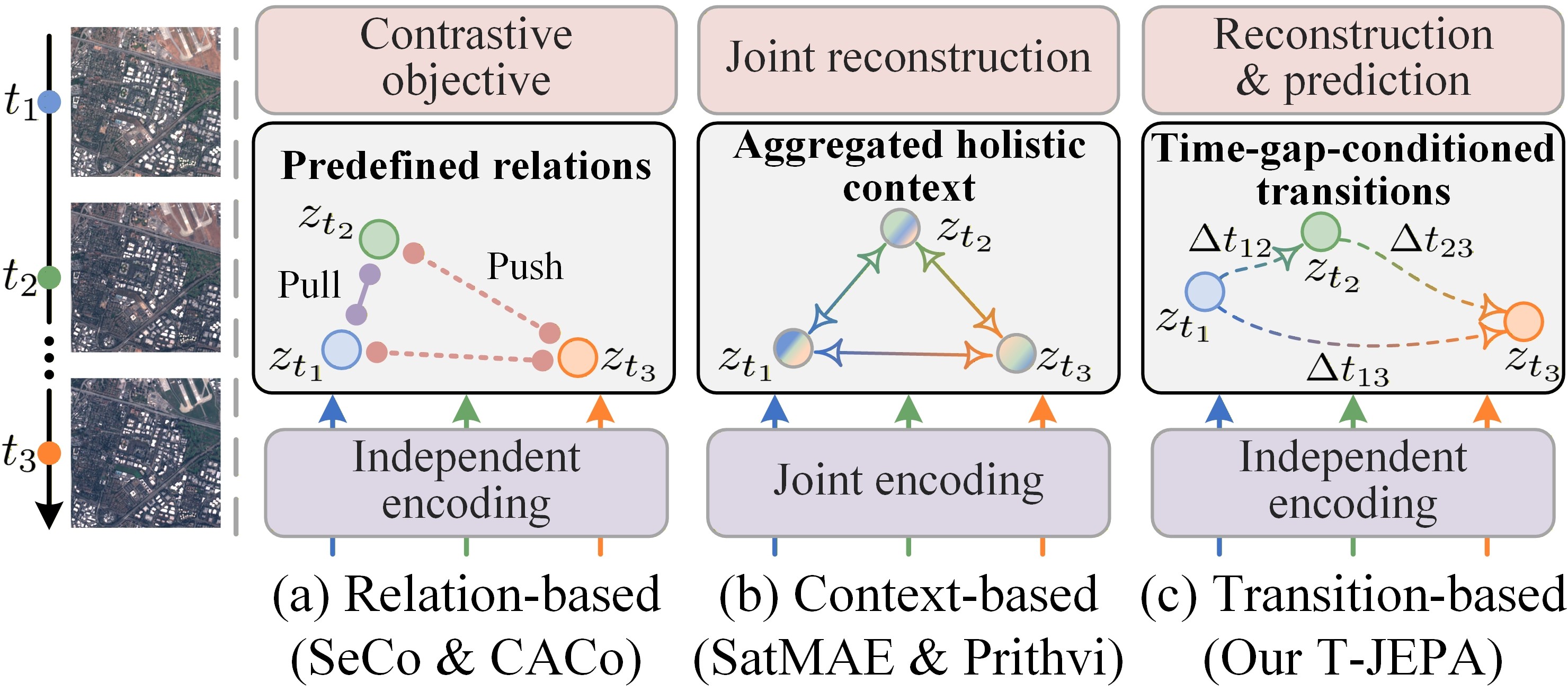}
				\caption{
					Temporal supervision paradigms for remote sensing representation learning. (a) Relation-based methods impose predefined pairwise constraints between observations. (b) Context-based methods jointly encode multiple observations to form holistic reconstruction context. (c) T-JEPA independently encodes each observation and learns time-gap-conditioned transitions between their representations.
				}
				\label{fig:overview}
			\end{figure}
			
			Existing remote sensing foundation models (RSFMs) mainly exploit sequential observations through relation-based or context-based paradigms, as illustrated in Figure~\ref{fig:overview}. Relation-based approaches, such as SeCo~\cite{seco} and CACo~\cite{caco}, use seasonal invariance or change-aware sampling to construct pairwise constraints. Context-based masked autoencoders (MAE), exemplified by SatMAE~\cite{satmae} and Prithvi-EO-2.0~\cite{prithvi}, aggregate cross-timestamp context for masked reconstruction. These paradigms resonate with two central ideas in video self-supervised learning (SSL): temporally related observations provide relational supervision~\cite{timecontrast}, while cross-time redundancy supports missing content recovery~\cite{maest,videomae}. Such intuitions are particularly natural for videos, whose neighboring frames are densely sampled, evolve smoothly, and exhibit substantial redundancy.

			EO sequences, however, operate in a different temporal regime. Acquisitions are typically sparse and nonuniform, so even consecutive observations may be separated by days, months, or years and describe evolution over markedly different temporal scales. Rather than treating this variability only as a complication, we regard the actual elapsed time as an additional supervisory signal that characterizes the scale over which one observed state evolves into another. Thus, we introduce {time-gap-conditioned predictability} as a representational inductive bias. Specifically, we seek source representations that, together with the elapsed time, retain sufficient information to infer the predictable aspects of the corresponding target state. This formulation does not assume fully deterministic Earth-surface evolution. Instead, it encourages the representation space to capture regular land-surface dynamics and organize observations according to their time-conditioned evolution.

            Directly realizing this predictive bias with pixel-space targets is ill-suited to EO sequences, because acquisition-specific factors such as clouds, illumination, and sensor effects, etc., introduce substantial uncertainty and instability into pixel-level supervision, obscuring the underlying land-surface dynamics. We therefore propose T-JEPA, a temporal joint-embedding predictive architecture (JEPA) that learns time-gap-conditioned transitions in latent space. By predicting target representations rather than future pixels~\cite{lecun2022path}, T-JEPA focuses supervision on the predictable components of land-surface evolution. Given a masked source latent representation and the elapsed time, a temporal predictor estimates the complete target latent field produced by an exponential-moving-average (EMA) encoder. Explicit time-gap conditioning distinguishes transitions over different durations and accommodates sparse and irregular sampling. In parallel, a masked-reconstruction~\cite{mae} branch preserves complementary spatial and appearance information in a separate decoder subspace.

			To promote reliable transition learning, T-JEPA directly supervises all temporally ordered observed pairs across multiple temporal scales instead of recursively rolling out intermediate predictions~\cite{vjepatwo}. This provides direct constraints on both short- and long-range transitions while avoiding error propagation. We further adopt asymmetric metadata injection: the online encoder receives spatiotemporal metadata with high dropout ratio, whereas the target encoder remains metadata-free. This prevents metadata from directly defining the latent prediction targets, reducing shortcut learning and reliance on auxiliary information.
			
			We fairly train and compare ten general SSLs and RSFMs under the same pre-training corpus and matched data throughput. T-JEPA consistently improves transfer across static and temporal downstream tasks, rivaling and often surpassing Prithvi-EO-2.0-TL~\cite{prithvi}. Beyond transfer performance, it induces a more temporally structured representation geometry, showing stronger cross-period consistency while preserving representation diversity and semantic discriminability, with transition predictability remaining strongly tied to the actual elapsed time. Decoded annual trajectories further reveal coherent seasonal evolution and plausible unobserved states, highlighting the potential of T-JEPA as a step toward remote sensing world models.
			
			Our contributions are summarized as follows:
			\begin{itemize}
				\item We identify the elapsed time between sparsely and irregularly sampled EO sequences as an underexploited source of supervision, and formulate time-gap-conditioned predictability as a representational inductive bias.
                \item We propose T-JEPA, which extends the JEPA paradigm to learn time-gap-conditioned latent transitions. It combines direct multi-scale transition supervision and asymmetric metadata injection with complementary masked reconstruction in a separate decoder subspace.
	            \item T-JEPA exhibits leading transfer performance across static and temporal tasks and learns representations with promising cross-period consistency, diversity, and discriminability, while exhibiting time-gap-dependent transition predictability and coherent latent trajectories.
			\end{itemize}
						
			\section{Related Works}
			
			\subsection{SSL for Temporal Visual Data}
			
			Videos provide rich self-supervision through dense sampling and strong short-term redundancy~\cite{slowfeat}. Classical objectives exploit temporal coherence~\cite{timecontrast,goroshin2015unsupervised} and future prediction~\cite{deeppred,advpred,vondrick2016anticipating}. Modern frameworks extend these ideas through temporal view matching~\cite{teeti2023temporal}, masked video modeling~\cite{videomae,maest}, and correspondence learning. In particular, SiamMAE~\cite{siammae} learns cross-frame correspondence through a Siamese architecture and asymmetric masking. These works establish invariance, reconstruction, and correspondence as effective temporal signals. EO archives share this sequential nature but exhibit sparse, irregular sampling and acquisition-dependent appearance variations.
			
			\subsection{Temporal Supervision in RSFMs}
			
			RSFMs exploit sequential observations through relation-based and context-based objectives. SeCo~\cite{seco} learns seasonal invariance, while CACo~\cite{caco} constructs change-aware contrastive supervision from short- and long-term differences. Context-based MAEs, such as SatMAE~\cite{satmae} and Prithvi-EO-2.0~\cite{prithvi}, encode acquisition time and aggregate cross-timestamp context for reconstruction. Together, these methods use temporal observations to define relations, provide metadata, or enrich reconstruction context. T-JEPA investigates a complementary signal: using the actual elapsed time to organize encoded EO states into temporal trajectories. Concurrent work Sat-JEPA-Diff~\cite{satjepadiff} predicts the embedding of the next satellite image to condition a latent diffusion model. Unlike T-JEPA, its primary goal is satellite image forecasting and generation rather than general-purpose transferable representation learning, and it does not explicitly condition latent transitions on variable elapsed intervals.
            
			\subsection{Predictive Learning in  Latent Space}
						
			Observation-space prediction is often complicated by noise, ambiguity, and incidental details. Recent works in time-series forecasting~\cite{latenttsf}, language modeling~\cite{nitp}, and video generation~\cite{pvvae} instead predict latent states or implicit tokens. Similarly, JEPAs \cite{lecun2022path,ijepa} predict target representations rather than reconstructing full observations. V-JEPA~\cite{vjepa,vjepatwo} predicts masked spatiotemporal representations, while LeWorldModel~\cite{leworldmodel} studies next-embedding prediction for physically structured dynamics. T-JEPA extends the JEPA principle to learn better representations from sparse, irregular EO sequences through continuous time-gap conditioned temporal prediction.
									
			\section{Method}

			\subsection{Overview}

            We propose T-JEPA to turn the sparse and nonuniform temporal sampling of EO sequences into explicit supervision for representation learning. Unlike spatiotemporal architectures that concatenate multi-date tokens and model their interactions as holistic context~\cite{videomae,vjepa,prithvi,a2mae,satmae}, T-JEPA independently encode each observation and learns time-gap-conditioned transitions between the encoded states. Temporal structure is therefore introduced through the pre-training objective rather than cross-time interaction in the backbone. This benefits both single-observation and multi-observation downstream tasks.

            Following the JEPA paradigm \cite{lecun2022path,ijepa}, T-JEPA implements temporal prediction in latent space, reducing the influence of uncertain and acquisition-specific appearance variations such as clouds and illumination. This formulation nevertheless raises two non-trivial challenges. First, injecting geospatial and temporal metadata into both prediction pathways creates shortcuts. Second, predictive supervision alone underemphasize fine-grained spatial and spectral details. T-JEPA addresses these issues through asymmetric metadata injection and a complementary masked-pixel reconstruction branch operating in a separate decoder subspace. A metadata-free EMA target encoder provides complete latent targets, while direct supervision over multiple source--target intervals covers transitions at different temporal scales. After pre-training, all auxiliary modules are discarded, leaving a standalone frame encoder for downstream transfer.
			
			\subsection{Frame-wise Encoding and Asymmetric Targets}
			
			Consider an ordered sequence $\{(x_{t_i},t_i)\}_{i=1}^{T}$ acquired at the same geographic location $\ell$, where $t_1<\cdots<t_T$. Each observation is divided into $L$ image patches, and $D$ denotes the latent dimension of the frame encoder. We maintain an online encoder $f_{\theta}$ and an EMA target encoder $f_{\bar{\theta}}$. 
			
			Spatiotemporal metadata awareness is important for remote sensing representation learning. Geographic coordinates and acquisition timestamps provide useful context to establish spatiotemporal correlations and are commonly encoded as additional embeddings \cite{satmae,prithvi}. However, using the same metadata on both sides of latent prediction introduces a shortcut. Since source and target observations share geographic cues and the timestamps can be inferred by time gap, the encoder and predictor may rely on metadata correspondence rather than the visual state of the source image. Such a shortcut can weaken the pressure to learn image-grounded representations.
			
			To address this issue, T-JEPA adopts \emph{asymmetric metadata injection}. The online pathway receives a masked observation together with optional location (latitude and longitude) and acquisition-time (year and day-of-year) metadata:
			\begin{equation}
				z_{t_i} = f_{\theta} \left(\mathcal{M}(x_{t_i});b_i e_{\mathrm{meta}}(\ell,t_i)\right),
				\quad
				z_{t_i}\in\mathbb{R}^{|\mathcal{V}_i|\times D},
				\label{eq:online_encoding}
			\end{equation}
            where $\mathcal{M}$ denotes random patch masking, $\mathcal{V}_i$ is the set of visible patch indices, and
			\begin{equation}
				e_{\mathrm{meta}}(\ell,t_i) = e_{\mathrm{loc}}(\ell) + e_{\mathrm{time}}(t_i)
			\end{equation}
			combines geographic and temporal embeddings. Both embeddings are implemented with sinusoidal encoding. Metadata dropout is controlled by
			\begin{equation}
				b_i\sim\mathrm{Bernoulli}(1-p_{\mathrm{meta}}).
			\end{equation}
			We use $p_{\mathrm{meta}}=0.5$, so the online encoder must also learn valid representations from image content alone. In contrast, the target encoder processes the complete observation without geographic or temporal metadata:
			\begin{equation}
				\bar z_{t_i} = \mathrm{sg} \left[f_{\bar{\theta}}(x_{t_i})\right],\quad \bar z_{t_i}\in\mathbb{R}^{L\times D},
				\label{eq:target_encoding}
			\end{equation}
			where $\mathrm{sg}(\cdot)$ denotes stop-gradient. The target encoder is updated as an EMA of the online visual encoder:
			\begin{equation}
				\bar{\theta} \leftarrow m\bar{\theta} + (1-m)\theta,
				\label{eq:ema_update}
			\end{equation}
			where $m$ is the momentum coefficient.
			
			This design provides three complementary effects. First, metadata cannot directly define the latent prediction targets, which encourages the predictor to rely on visual state transitions. Second, metadata dropout regularly forces the online encoder to learn image-grounded features through both reconstruction and temporal prediction, which in turn improves the visual representations transferred to the target encoder. Third, the resulting online encoder remains robust when metadata are unavailable in downstream applications.
			
			\begin{figure}[tb]
				\centering
				\includegraphics[width=\linewidth]{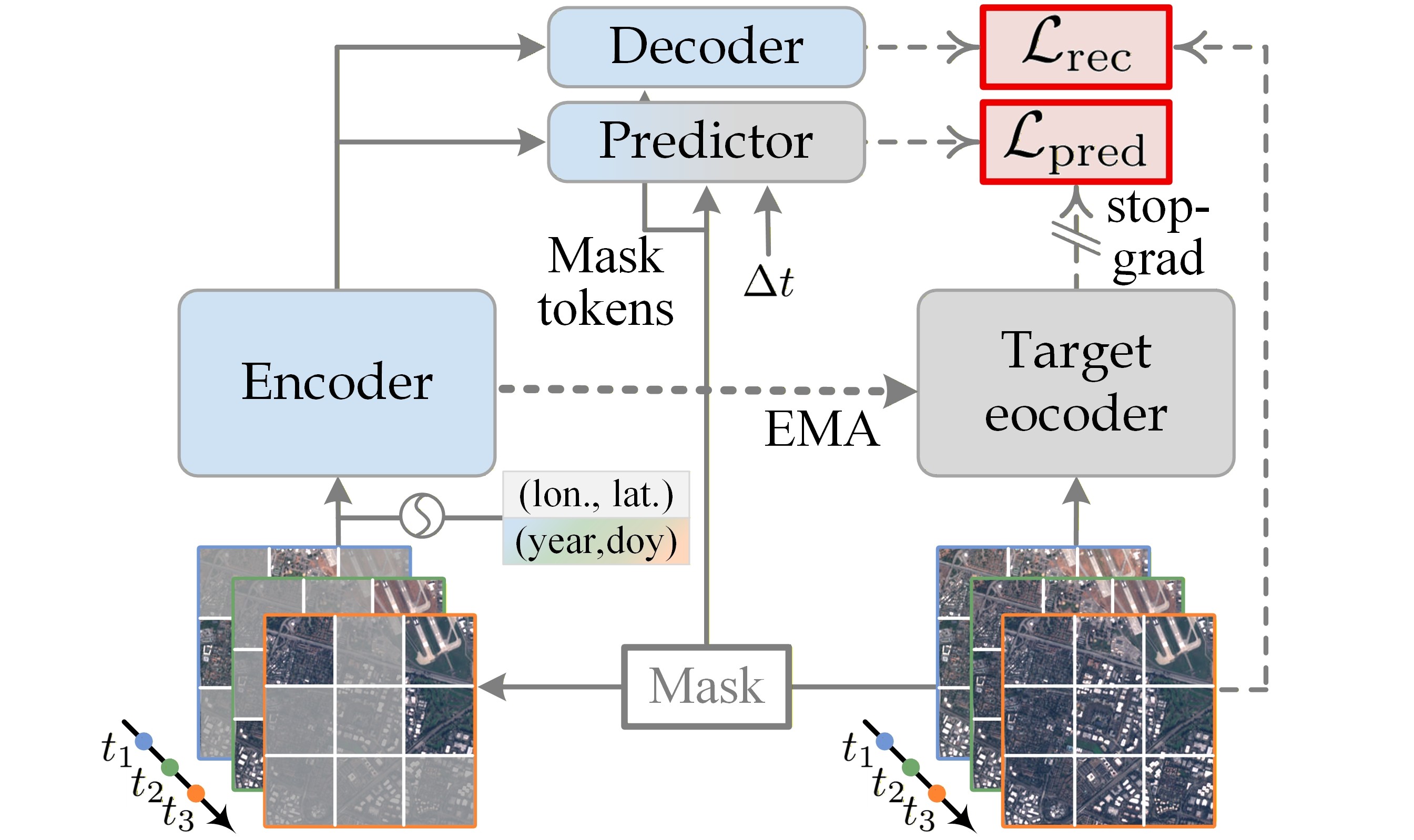}
				\caption{
					Overview of T-JEPA. A shared online encoder independently processes masked observations with optional spatiotemporal metadata. The reconstruction branch predicts masked image patches, while the time-gap-conditioned joint-embedding branch predicts the complete latent field of a target observation produced by a metadata-free EMA encoder. Gradients are stopped through the target pathway, and all auxiliary modules are discarded after pre-training.
				}
				\label{fig:method}
			\end{figure}
			
			\subsection{Time-Gap-Conditioned Temporal Prediction}
			EOs are sparsely and irregularly sampled, so transitions spanning days, months, or years should not be treated as equivalent. For a source--target pair $(i,j)$ with $i<j$, we define the elapsed time (in days) as
			\begin{equation}
				\Delta t_{ij} = t_j-t_i.
				\label{eq:time_gap}
			\end{equation}
			A temporal condition embedder maps this interval into a conditioning vector:
			\begin{equation}
				c_{ij} = e_{\Delta} \left(\Delta t_{ij}\right),
				\label{eq:time_embedding}
			\end{equation}
            where $e_{\Delta}$ is implemented with sinusoidal encoding followed by a two-layer MLP to encode time gaps.
		
			Before temporal prediction, learned mask tokens are inserted to restore the complete spatial token layout. A temporal predictor $g_{\phi}$ then estimates the complete target latent:
			\begin{equation}
				\hat z_{t_j|t_i}=g_{\phi}\left(z_{t_i};c_{ij}\right), \quad \hat z_{t_j|t_i}\in\mathbb{R}^{L\times D}.
				\label{eq:temporal_prediction}
			\end{equation}
			Following adaptive Layer Normalization (AdaLN)~\cite{dit}, the condition $c_{ij}$ generates modulation parameters for the Transformer blocks in $g_{\phi}$. The predictor therefore adapts its transformation to the actual duration between the source and target states.

			Unlike masked pixel reconstruction, which is supervised only on removed image patches, temporal prediction is supervised against the complete target latent field:
			\begin{equation}
				\mathcal{L}_{i\rightarrow j} = \frac{1}{L} \sum_{p=1}^{L}\left\|\hat z_{t_j|t_i,p}-\bar z_{{t_j},p}\right\|_2^2.
				\label{eq:pair_loss}
			\end{equation}
			Predicting representations rather than future pixels avoids requiring the model to reproduce acquisition-dependent variations such as clouds, illumination, atmospheric conditions, and sensor effects. Instead, the objective encourages the source representation to retain information that supports prediction of the target latent state over the specified time gap.
            
			\paragraph{Direct multi-scale transition supervision.}
			
			To cover the diverse temporal intervals within an EO sequence, T-JEPA directly constructs all forward source--target pairs:
			\begin{equation}
				\mathcal{P} = \left\{(i,j)\mid 1\leq i<j\leq T\right\}.
				\label{eq:temporal_pairs}
			\end{equation}
			The temporal prediction objective is
			\begin{equation}
				\mathcal{L}_{\mathrm{pred}} = \frac{1}{|\mathcal{P}|} \sum_{(i,j)\in\mathcal{P}} \mathcal{L}_{i\rightarrow j}.
				\label{eq:prediction_loss}
			\end{equation}
			This formulation directly supervises both short- and long-range transitions. Unlike recursive rollout~\cite{vjepatwo}, every target is predicted from an observed source representation rather than an intermediate predicted state, avoiding error propagation while retaining direct supervision at multiple temporal scales.
			
			\paragraph{Complementary reconstruction objective.}
			
			The reconstruction branch follows the standard masked pixel reconstruction objective of MAE~\cite{mae}, denoted by $\mathcal{L}_{\mathrm{rec}}$. It uses a decoder space independent of the temporal predictor. The final pre-training objective is
			\begin{equation}
				\mathcal{L} = \mathcal{L}_{\mathrm{rec}} + \lambda\mathcal{L}_{\mathrm{pred}},
				\label{eq:overall_loss}
			\end{equation}
			where $\lambda$ controls the strength of temporal predictive supervision. Masked reconstruction preserves fine-grained spatial, spectral, and appearance information, while joint-embedding prediction structures its cross-time evolution in latent space.
		
			\begin{table*}[tbp]
				\centering
				\begin{tabular}{l|cccc|cccccc|ccc}
					& \multicolumn{4}{c|}{Linear classification} & \multicolumn{6}{c|}{Semantic segmentation} & \multicolumn{3}{c}{Change detection} \\
					& \multicolumn{2}{c}{BigEarthNet} & \multicolumn{2}{c|}{PASTIS} & \multicolumn{3}{c}{DFC20 (no metadata)} & \multicolumn{3}{c|}{PASTIS ($T=8$)} & \multicolumn{3}{c}{OSCD} \\
					Method & $1\%$ & $10\%$ & $T=1$ & $T=8$ & OA & $\kappa$ & mIoU & OA & $\kappa$ & mIoU & Prec. & Rec. & F1 \\ \Xhline{1pt}
					Random init. & 37.8 & 41.5 & 22.7 & 29.2 & 58.4 & 49.4 & 31.8 & 79.5 & 74.4 & 46.6 & 47.1 & 55.9 & 51.1 \\
					MAE & 58.3 & 63.8 & 43.0 & 45.9 & 64.8 & 57.3 & 37.6 & 84.0 & 80.1 & 54.7 & 66.0 & 50.5 & 57.3 \\
					I-JEPA & 59.1 & 65.1 & 40.3 & 44.9 & 65.5 & 58.3 & 38.9 & 84.6 & 80.9 & 56.3 & 59.2 & 54.2 & 56.6 \\
					V-JEPA & 52.6 & 58.0 & 33.7 & 41.6 & 63.2 & 55.6 & 36.9 & 84.4 & 80.5 & 56.8 & 56.4 & 57.2 & 56.8 \\
					SiamMAE & 59.3 & 66.5 & 43.0 & 46.1 & 64.2 & 56.6 & 37.2 & 84.1 & 80.3 & 55.1 & 60.1 & \textbf{59.0} & 59.5 \\
					SeCo & 46.2 & 50.8 & 32.4 & 39.5 & 57.4 & 48.1 & 30.4 & 77.7 & 72.3 & 44.0 & 56.7 & 47.9 & 51.9 \\
					CACo & 45.7 & 50.5 & 30.1  & 38.5 & 56.5 & 47.3 & 30.6 & 77.8 & 72.4 & 44.5 & 51.7 & 52.9 & 52.3 \\
					SeaMo & 59.5 & 64.5 & 44.9 & 46.8 & 64.7 & 57.3 & 38.1 & 84.1 & 80.2 & 54.9 & 58.9 & 57.1 & 58.0 \\
					SatMAE & 61.1 & 67.7 & 44.1 & 48.5 & 63.3 & 55.7 & 36.8 & 84.9 & 81.2 & 57.1 & 59.4 & 55.7 & 57.5 \\
					Prithvi & 62.4 & 69.6 & 46.6 & 49.8 & 63.9 & 56.4 & 37.1 & 85.3 & 81.8 & 57.9 & 61.6 & 58.4 & 59.9 \\
					\textbf{T-JEPA} & \textbf{66.0} & \textbf{72.1} & \textbf{46.8} & \textbf{50.7} & \textbf{66.5} & \textbf{59.2} & \textbf{39.4} & \textbf{85.4} & \textbf{81.9} & \textbf{58.4} & \textbf{67.0} & 54.8 & \textbf{60.3}  \\
					
				\end{tabular}
				\caption{Downstream task performance. Metrics include macro mean average precision (mAP) for linear classification; overall accuracy (OA), Cohen's kappa 	($\kappa$), and mean intersection over union (mIoU) for semantic segmentation; and precision, recall, and F1 score for change detection. The best results are in \textbf{bold}.}
				\label{tab:downstream}
			\end{table*}
			
			\section{Experiments}
			
			\subsection{Experimental Setting}
			
			\paragraph{Pre-training data and comparison protocol.}
            We evaluate ten methods in total: T-JEPA, MAE~\cite{mae}, I-JEPA~\cite{ijepa}, V-JEPA~\cite{vjepa}, SiamMAE~\cite{siammae}, SeCo~\cite{seco}, CACo~\cite{caco}, SeaMo~\cite{seamo}, SatMAE~\cite{satmae}, and Prithvi-EO-2.0-TL (Prithvi)~\cite{prithvi}. Among them, SatMAE, Prithvi, and our T-JEPA utilize acquisition metadata. To control for the substantial differences among published RSFMs~\cite{nooneknows}, all methods are implemented with the same HiViT-S/16 backbone~\cite{hivit}, pre-trained on the CACo dataset with equivalent data throughput. The CACo dataset contains ten Sentinel-2 RGB observations from each of 200K globally distributed locations, with half acquired during 2017--2018 and the other during 2021--2022. 
			
			\paragraph{Implementation details.}
			T-JEPA sees four random observations per location at each iteration, is trained for 100 epochs on eight RTX-4090D GPUs with a global batch size of 1,024 (locations) and a base learning rate of $1.5\times10^{-5}$. Baseline training durations are adjusted according to their temporal input sizes to match the image throughput of T-JEPA. T-JEPA's reconstruction and prediction decoders have matched widths and depths. The EMA momentum increases linearly from 0.998 to 0.9996~\cite{vjepa}; $\lambda$ is set to 0.5. 
					
			\begin{figure*}[tb]
				\centering
				\includegraphics[width=\linewidth]{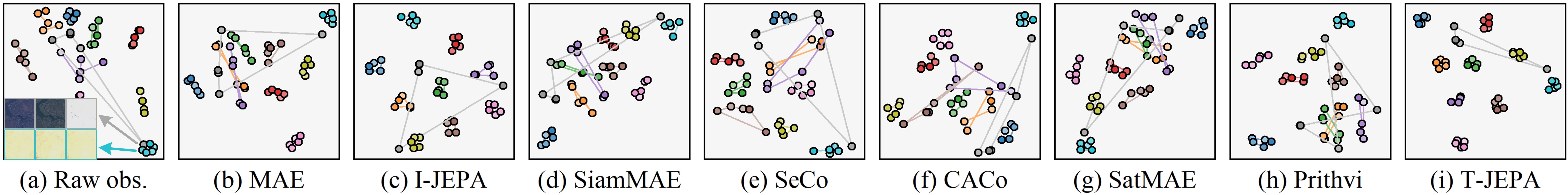}
				\caption{t-SNE visualization of raw observations and learned representations. We randomly select 10 CACo locations, each with five observations acquired during 2017--2018. Panel (a) shows raw observations, and the remaining panels show representations extracted by the compared pre-trained models. T-JEPA preserves the natural location-wise organization while forming more compact clusters with structured intra-location variation.}
				\label{fig:tsne}
			\end{figure*}
			
			\subsection{Downstream Performance}
			
			Table~\ref{tab:downstream} evaluates T-JEPA under two complementary downstream input regimes. Single-observation (static) transfer examines whether temporal pre-training improves the standalone frame encoder, whereas multi-observation (temporal) transfer evaluates whether independently extracted representations can be effectively integrated across time. All methods use the same backbone and downstream protocols, and both pre-training and transfer use only Sentinel-2 RGB imagery. Additional dataset details and evaluation protocol are provided in the supplement.
			
			\paragraph{Static transfer.}
			We evaluate linear classification on BigEarthNet~\cite{ben} under the $1\%$ and $10\%$ labeled-data settings and on PASTIS~\cite{pastis} with a single observation ($T=1$). We further fine-tune the encoder with UperNet~\cite{upernet} for semantic segmentation on DFC20~\cite{robinson2021global}. Notably, DFC20 provides no acquisition metadata, testing robustness to a mismatch with metadata-aware pre-training of SatMAE, Prithvi, and our T-JEPA.

			T-JEPA achieves the best results across all single-observation settings. It reaches 66.0 and 72.1 mAP on BigEarthNet, outperforming Prithvi by 3.6 and 2.5 points, respectively, and obtains 46.8 mAP on single-frame PASTIS. On DFC20, T-JEPA achieves 66.5 OA, 59.2 $\kappa$, and 39.4 mIoU, despite the absence of downstream metadata. These results show that temporal predictive supervision between independently encoded states improves visual representations without making the encoder dependent on multi-temporal inputs or auxiliary metadata.

			\paragraph{Temporal transfer.}			
			We evaluate multi-temporal land-cover understanding on PASTIS using eight observations ($T=8$) for both linear classification and semantic segmentation. We also evaluate bi-temporal change detection on OSCD~\cite{oscd}. To isolate the quality of the learned representations, all methods encode each observation independently. For PASTIS, the frame-wise features are temporally averaged before being passed to the task head; for OSCD, the feature difference between the two observations is used by the UperNet.
			
            T-JEPA reaches 50.7 mAP for PASTIS classification and 58.4 mIoU for PASTIS segmentation, achieving the best performance in both settings. On OSCD, it obtains the best F1 score of 60.3. Together, these results show that its single-frame representations remain individually discriminative and can also be effectively combined for temporal aggregation and cross-date change reasoning.
			
			\subsection{Representation Geometry and Temporal Dynamics}
			
			\paragraph{Geometry of learned representations.}
			
			EO sequences possess an intrinsic structure: repeated acquisitions from the same location share persistent land-cover and geographic characteristics, while seasonal and acquisition-dependent variations create substantial intra-location dispersion. Since T-JEPA aims to exploit this structure as temporal supervision, we first examine whether it is preserved and more clearly organized in the learned representation space.

            \begin{table}[htb]
                \centering
                \setlength{\tabcolsep}{4pt}
                \begin{tabular}{l|ccccc}
                    Method  & CKA & R$@$1  & R$@$5 & ER & mAP$\uparrow$ \\ \Xhline{1pt}
                    MAE & 0.52 & 48.31 & 54.75  & 134.04 & 49.14 \\
                    I-JEPA & 0.21 & 88.78 & 93.42 & 233.06 & 50.96 \\
                    V-JEPA & 0.19 & 43.82 & 58.75 & 156.68 & 44.38 \\
                    SiamMAE & 0.60 &  37.21 & 47.71  & 132.02 & 55.15 \\
                    SeCo & 0.34 &  31.77 & 49.35  & 25.69 & 45.72 \\
                    CACo & 0.40 &  31.35 & 47.16  & 26.28 & 46.18 \\
                    SeaMo & 0.60 & 44.88 & 52.30  & 136.85 & 51.13 \\
                    SatMAE & 0.71 & 42.51 & 52.58  & 125.67 & 52.37 \\
                    Prithvi & 0.66 &  42.97 & 53.22  & 134.76 & 53.76 \\
                    \textbf{T-JEPA}  & 0.65 &  81.84 & 93.32 & 229.91 & \textbf{56.92} \\
                \end{tabular}
                
                \caption{CKA measures global alignment with the raw-observation geometry; R$@$1 and R$@$5 evaluate local cross-period consistency through retrieval among 2,000 locations; Effective Rank (ER) diagnoses representation diversity and potential rank collapse; and mAP reports semantic discriminability using $k$NN classification ($k=20$) on BigEarthNet ($1\%$). These metrics are interpreted jointly, since neither higher CKA nor higher ER alone necessarily indicates a better representation.}
                \label{tab:geometry}
            \end{table}

            Figure~\ref{fig:tsne} visualizes raw observations and representations extracted by different pre-trained encoders. Raw observations already exhibit a location-wise organization, but their clusters are dispersed by seasonal and acquisition-dependent appearance changes. As highlighted in the figure, an overexposed observation from one location is embedded near another location whose observations exhibit similarly high brightness, illustrating how acquisition-specific appearance can obscure the geographic organization. T-JEPA produces more compact location-wise groupings while retaining structured variation among observations from the same location, suggesting that it strengthens the intrinsic organization of EO data rather than merely enforcing uniform invariance.

            \begin{figure*}[tb]
            \centering
            \includegraphics[width=\linewidth]{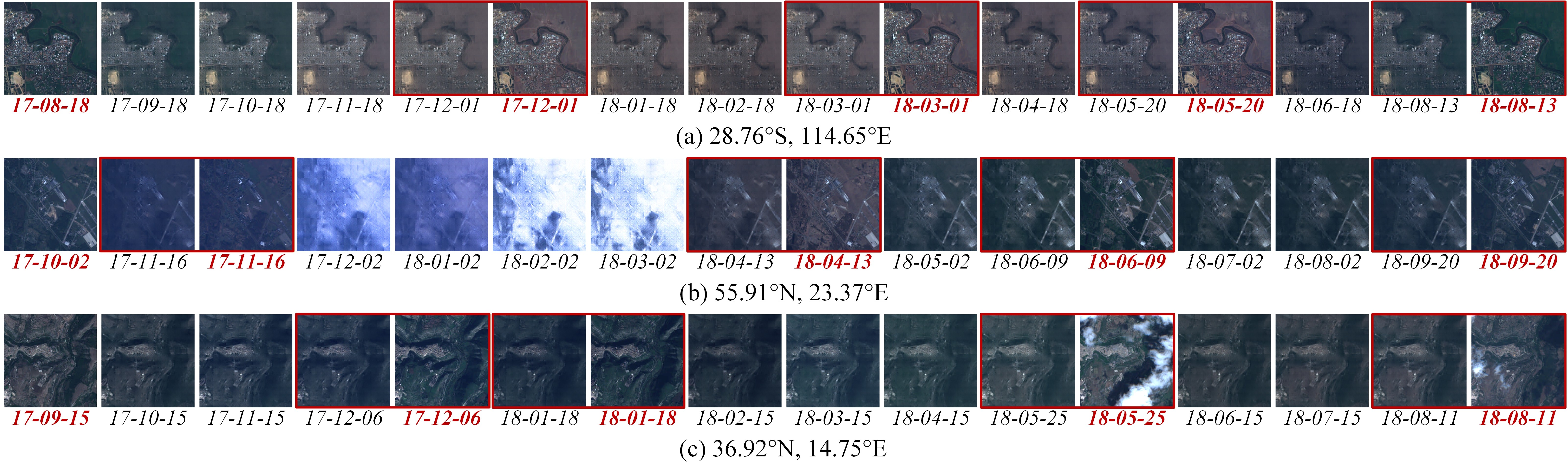}
            \caption{
                Visualization of temporal dynamics learned by T-JEPA. Starting from the anchor observation in the first column, the model predicts and decodes a full annual trajectory. Ground-truth observations are indicated by dates shown in bold red. The predicted states generally follow the seasonal vegetation trends in the available observations and reveal plausible unobserved states, such as continuous winter snow cover. Predictions at cloud-contaminated dates often preserve a clear underlying land-surface state.}
            \label{fig:pred}
            \end{figure*}

            We further quantify this geometry from four complementary perspectives; please see Table~\ref{tab:geometry}. T-JEPA maintains substantial alignment with the raw-observation geometry, achieving a CKA of 0.65 comparable to SatMAE and Prithvi, while attaining near-best cross-period retrieval. It also retains a high effective rank of 229.91 and achieves the best $k$NN mAP of 56.92. Together, these results indicate that T-JEPA preserves the global structure of observations while reorganizing it into a more coherent location-wise geometry with high diversity and semantic discriminability.

			\paragraph{Time-conditioned transition predictability.}
			Location consistency alone does not establish that representations encode temporal transitions, since observations from the same location may remain close without distinguishing how their states evolve over time. We therefore freeze each pre-trained encoder and train the same lightweight two-layer MLP on all forward temporal pairs. The pooled source representation is concatenated with a time-gap embedding and mapped to the target representation. This probe is independent of T-JEPA's pre-training predictor and is identically applied to all methods; details are provided in the supplement.
			
			T-JEPA achieves a Target R$@$1 of 87.50\%, substantially exceeding SatMAE (68.75\%), V-JEPA (63.95\%), and Prithvi (63.90\%). Shuffling the time interval reduces its retrieval accuracy to 30.00\%, the largest degradation among the compared methods. This result distinguishes time-conditioned predictability from generic location consistency. For example, I-JEPA achieves the highest cross-period location retrieval in Table~\ref{tab:geometry}, yet obtains only 7.80\% Target R$@$1 in the transition probe. T-JEPA therefore does not merely bring repeated observations of the same location closer; it organizes them into distinguishable transitions that are strongly associated with the actual elapsed time.
			
			\begin{table}[tbp]
				\centering
				\setlength{\tabcolsep}{4pt}
				\begin{tabular}{l|cccc}
					Method  & PredSim & Gain & \makecell[c]{Target R$@$1$\uparrow$} & \makecell[c]{Shuffled R$@$1} \\ \Xhline{1pt}
					MAE & 0.94 & 0.05 & 24.25 & 18.75 \\    
					I-JEPA & 0.77 & 0.02 & 7.80 & 7.40\\
					V-JEPA & 0.91 & 0.12 & 63.95 & 33.50 \\
					SiamMAE & 0.99 & 0.01 & 30.90 & 19.95 \\
					SeCo & 0.82 & 0.07 & 11.90 & 9.85 \\
					CACo & 0.86 & 0.05 & 14.75 & 12.10 \\
					SeaMo & 0.92 & 0.07 & 25.85 & 19.00 \\
					SatMAE & 0.96 & 0.07 & 68.75 & 28.85 \\
					Prithvi & 0.98 & 0.03 & 63.90 & 25.25 \\
					\textbf{T-JEPA}  & 0.95 & 0.08 & \textbf{87.50} & 30.00 \\
				\end{tabular}
				\caption{Time-conditioned predictability of frozen representations. PredSim measures prediction--target similarity, Gain measures improvement over directly using the source representation, and Target R$@$1 retrieves the correct target among five observations of the same location. Shuffled R$@$1 replaces the true time interval with a shuffled one at inference.}
				\label{tab:predictability}
			\end{table}

            \begin{table*}[tbp]
            \centering
            \small
            \renewcommand{\arraystretch}{1.05}
            
            \begin{minipage}[b]{0.485\textwidth}
                \centering
                \setlength{\tabcolsep}{5pt}
                \begin{tabular}{cc|ccc}
                    Meta. & Pred. & BEN & PASTIS & DFC20 \\
                    \Xhline{1pt}
                    \multicolumn{2}{c|}{\textcolor{gray}{\textit{MAE baseline}}} & 58.3 & 45.9 & 37.6 \\
                    & $\checkmark$ & \textbf{62.4} & \textbf{47.6} & \textbf{38.8} \\ \hline
                    $\checkmark$ &  & 60.5 & 46.8 & 38.3 \\  
                    $\checkmark$ & $\checkmark$
                    & \textbf{66.0} & \textbf{50.7} & \textbf{39.4} \\
                \end{tabular}\\
                \vspace{2pt}{(a) Component contributions}
            \end{minipage}
            \hfill
            \begin{minipage}[b]{0.485\textwidth}
                \centering
                \setlength{\tabcolsep}{5pt}
                \begin{tabular}{cc|ccc}
                    Online meta. drop &
                    Target meta. &
                    BEN & PASTIS & DFC20 \\
                    \Xhline{1pt}
                    0   & $\checkmark$ & 62.9 & 48.9 & 37.6 \\
                    0   &            & 65.1 & 50.4 & 38.5 \\
                    0.5 &            &
                    \textbf{66.0} & \textbf{50.7} & \textbf{39.4} \\
                \end{tabular}\\
                \vspace{2pt}{(b) Asymmetric metadata injection}
            \end{minipage}
            
            \vspace{0.8em}
            
            \begin{minipage}[t]{0.485\textwidth}
                \centering
                \setlength{\tabcolsep}{5pt}
                \begin{tabular}{cc|ccc}
                    Recon. target &
                    Pred. target &
                    BEN & PASTIS & DFC20 \\
                    \Xhline{1pt}
                    pixel  & pixel  & 64.3 & 49.5 & 37.8 \\
                    latent & latent & 61.7 & 48.6 & 36.1 \\
                    pixel  & latent &
                    \textbf{66.0} & \textbf{50.7} & \textbf{39.4} \\
                \end{tabular}\\
                \vspace{2pt}{(c) Objective target spaces}
            \end{minipage}
            \hfill
            \begin{minipage}[t]{0.485\textwidth}
                \centering
                \setlength{\tabcolsep}{5pt}
                \begin{tabular}{l|ccc}
                    Prediction strategy & BEN & PASTIS & DFC20 \\  
                    \Xhline{1pt}
                    Rollout (recursive prediction) & 64.3 & 47.0 & 38.7 \\
                    Adjacent-pair (local transition) & 65.9 & 50.1 & 38.6 \\
                    All-pair (multi-scale transition) & \textbf{66.0} & \textbf{50.7} & \textbf{39.4} \\
                \end{tabular}\\
                \vspace{2pt}{(d) Temporal supervision strategy}
            \end{minipage}
            
            \caption{
                Ablation studies of the temporal prediction branch. We report linear classification mAP on BigEarthNet with $1\%$ labels and PASTIS with $T=8$, and mIoU of DFC20 semantic segmentation.
            }
            \label{tab:ablation}
			\end{table*}
			
			\paragraph{Emergent temporal dynamics.} 		
			To qualitatively examine what these latent transitions represent, we decode a full annual trajectory predicted from a single anchor observation. Given the anchor representation and twelve target time intervals, the model predicts monthly latent states without observing any future frame. We use a visualization-oriented variant that permits direct decoding of latent states.
						
			As shown in Figure~\ref{fig:pred}, the decoded trajectories exhibit coherent greening, senescence, and surface-color transitions that broadly follow the trends observed in the sparsely sampled ground-truth images. In some locations, T-JEPA predicts continuous winter snow cover even though none of the training observations contain visible snow, suggesting that it has learned geographically conditioned seasonal regularities rather than merely interpolating between observed frames.

			Predictions corresponding to cloud-contaminated observations also often remain cloud-free while preserving the underlying surface structure. This behavior is consistent with the latent objective focusing on predictable land-surface states rather than transient acquisition conditions. These qualitative results do not establish T-JEPA as a generative forecasting or cloud-removal system, but, together with the frozen-representation probe, they highlight the potential of time-conditioned latent prediction for all-weather surface-state inference and remote-sensing world modeling.

		    \subsection{Ablation Study}
						
			We progressively examine the temporal prediction branch in terms of component contributions, metadata injection, objective spaces, and prediction strategies. All results are reported in Table~\ref{tab:ablation}. We provide details about its computation overhead and loss weight in the supplement.

    		\paragraph{Component contributions.}
    		Temporal prediction alone provides larger and more consistent gains than metadata alone, while their combination performs best. This indicates that predictive learning is the primary source of improvement, with metadata serving as supporting context.
    		
            \paragraph{Asymmetric metadata injection.}
            Starting from the symmetric baseline that provides metadata to both encoders, removing metadata from the target encoder consistently improves all three benchmarks, supporting our concern that metadata-conditioned targets may create shortcuts. Further applying metadata dropout to the online encoder yields the best performance, encouraging content-grounded representations while retaining metadata as optional context.
    		
    		\paragraph{Objective target spaces.}
    		With independent task-specific decoders, pixel reconstruction and latent temporal prediction consistently outperform using the same target space for both objectives. Pixel reconstruction preserves fine-grained spatial and spectral information, whereas latent prediction structures cross-time transitions without modeling acquisition-specific pixel variations.
    		
    		\paragraph{Temporal supervision strategy.}
    		All-pair prediction performs best across the three settings. Compared with adjacent-pair supervision, it directly covers both short- and long-range transitions; compared with rollout, it predicts every target from an observed source and avoids error propagation through intermediate predictions.
    		
    		\paragraph{Prediction loss weight.}
    		A moderate prediction weight performs best, with $\lambda=0.5$ achieving the best overall results. Smaller weights underuse temporal supervision, whereas larger weights slightly weaken transfer, possibly because excessive predictability imposes an overly smooth constraint that suppresses fine-grained observation details.
    		
    		\paragraph{Computational overhead.} 
            T-JEPA increases forward computation by 220.90\% over MAE, of which the gradient-free EMA target-encoder forward accounts for 72.40\% of the added FLOPs and the prediction Transformer for 27.16\%; all remaining prediction components contribute less than 0.5\%. Although GPU-only training time increases by 88.08\%, the end-to-end epoch wall-clock time rises by only 20.00\%.
    			
    		\section{Conclusion}
            We introduced T-JEPA, a temporal JEPA that exploits time-conditioned transitions for remote sensing representation learning. By using elapsed intervals to supervise transitions between independent states, T-JEPA exploits sparse and irregular EO sampling without imposing multi-temporal dependence on the deployable encoder. Experiments demonstrated consistent gains across static and temporal tasks. Representation analyses further revealed time-conditioned predictability, cross-period consistency, semantic discriminability, and coherent latent dynamics. These findings establish T-JEPA as a promising step towards learning transferable remote sensing representations and modeling Earth-surface evolution.
            
            \section{Ethical Statement}
    
            T-JEPA may benefit environmental monitoring, land-cover mapping, and disaster assessment. However, stronger remote sensing representations could also facilitate intrusive surveillance, particularly when transferred to higher-resolution imagery. Moreover, predicted or decoded latent states should not be treated as actual observations or reliable forecasts. Responsible deployment should therefore ensure clear provenance, appropriate uncertainty communication, validation against real observations, and consideration of geographic bias, access control, and pre-training costs.
            
    		\bibliography{aaai2027}

\section{Supplementary Material}

\setcounter{figure}{0}
\setcounter{table}{0}
\setcounter{equation}{0}

\renewcommand{\thefigure}{S\arabic{figure}}
\renewcommand{\thetable}{S\arabic{table}}
\renewcommand{\theequation}{S\arabic{equation}}

This supplementary material provides additional dataset descriptions, implementation details, and extended experimental results for the main paper. For implementation details not explicitly covered here, please refer to our code.

\subsection{Implementation Details}

\paragraph{Land cover classification.}

We evaluate land-cover classification on BigEarthNet~\cite{ben} and PASTIS~\cite{pastis}. BigEarthNet contains approximately 590K Sentinel-2 image patches of size $120\times120$, with multi-label annotations over 19 land-cover classes. We report results using $1\%$ and $10\%$ of the training data. PASTIS contains 2,433 Sentinel-2 time series of size $128\times128$, with sequence lengths ranging from 38 to 61. Folds 1--4 are used for training and fold 5 for testing. We evaluate either the first observation from each sequence ($T=1$) or eight observations sampled at approximately uniform temporal intervals ($T=8$). Following~\cite{swinssl}, we convert the semantic segmentation maps into multi-label annotations by retaining crop categories whose area ratios exceed $10\%$. This results in 18 land-cover labels after excluding the \textit{Background} and \textit{Void label} categories.

For all classification evaluations, the pre-trained encoder is frozen and the input images are resized to $224\times224$ to match the pre-training resolution. Batch normalization followed by a single linear layer is applied to the frozen representations. The classifier is trained using binary cross-entropy loss and the Adam optimizer for 50 epochs with a batch size of 32. The initial learning rate is $10^{-3}$ and is reduced by a factor of 0.1 at epochs 30 and 40.

\begin{figure*}[tb]
	\centering
	\includegraphics[width=\linewidth]{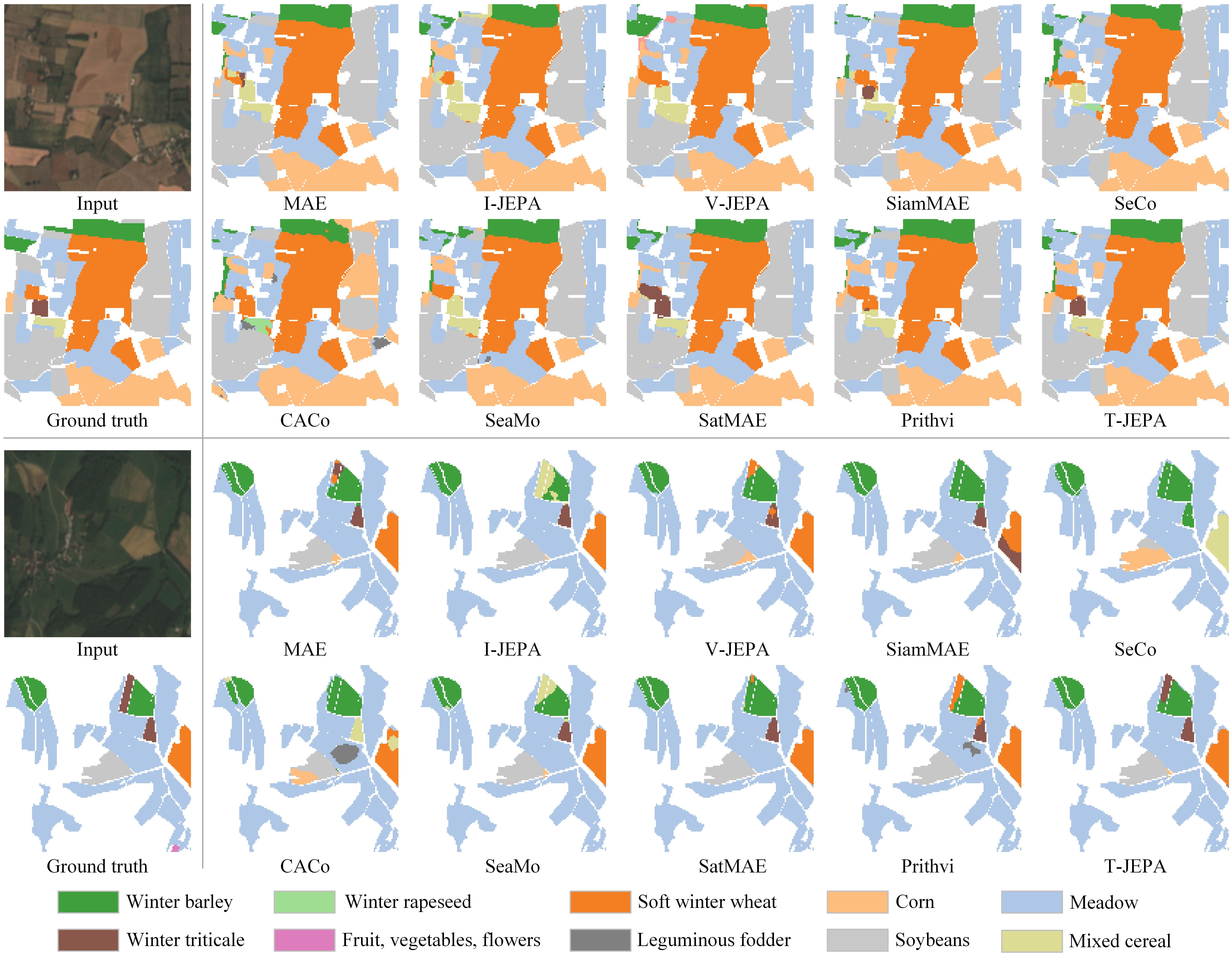}
    \caption{Qualitative comparison on PASTIS ($T=8$) segmentation.}
    \label{fig:pastis}
\end{figure*}

\paragraph{Semantic segmentation.}

We evaluate semantic segmentation on PASTIS with eight observations ($T=8$) and on DFC20~\cite{robinson2021global}. DFC20 contains 6,114 images of size $256\times256$ covering eight semantic classes, including 986 training images and 5,128 test images. We do not resize images on this dataset. Notably, DFC20 provides no acquisition metadata, creating a transfer mismatch for metadata-aware pre-training methods such as SatMAE, Prithvi, and T-JEPA. In this setting, the models must operate without the metadata cues available during pre-training.

We use UperNet~\cite{upernet} as the segmentation head and optimize the entire network using cross-entropy loss and AdamW for 50 epochs with a batch size of 8. The learning rate is linearly warmed up for five epochs to a maximum of $10^{-3}$ and then decayed to $10^{-6}$ following a cosine schedule. The weight decay is set to $10^{-2}$, and a layer-wise learning-rate decay of 0.9 is applied to the encoder. For PASTIS, the eight observations are encoded frame by frame, and their feature maps are temporally averaged before being passed to the segmentation head.

\paragraph{Change detection.}

We evaluate bi-temporal change detection on OSCD~\cite{oscd}, which contains14 image pairs for training and 10 pairs for testing. Following~\cite{seco}, the original images of varying spatial sizes are cropped into non-overlapping $96\times96$ patches. Each patch is resized to $224\times224$ before being processed by the encoder. The two observations are encoded separately, and their feature difference is passed to UperNet for binary change prediction.

To mitigate the severe class imbalance between changed and unchanged pixels, we combine binary cross-entropy loss with Dice loss~\cite{sudre2017generalised}. The maximum learning rate is $5\times10^{-4}$. The remaining optimization settings follow those used for semantic segmentation.

\paragraph{Frozen representation probe.}

We conduct the frozen representation probe on a 2,000-location subset of CACo. Although these locations are part of the SSL pre-training distribution for all compared methods, the goal of this experiment is to analyze the temporal structure encoded in their representation space rather than unseen-data generalization. For each location, we use the five chronologically ordered observations (2017--2018) and extract one L2-normalized global representation per timestamp using the frozen downstream encoder. All forward temporal pairs $(i,j)$ with $i<j$ are constructed, yielding ${5 \choose 2}=10$ transitions per location. The 2,000 locations are split into 1,600/200/200 locations for probe training, validation, and testing, respectively, ensuring that the newly trained predictor does not observe the same locations across its own splits.

The same conditional MLP predictor is trained independently for every pre-trained representation. The source representation is first normalized by LayerNorm. The temporal interval $\Delta t$ is converted to days, transformed using $\log(1+\Delta t)$, normalized by the maximum interval in the training split, and embedded by a two-layer MLP with a 64-dimensional output. The source representation and time embedding are concatenated and passed through an MLP with two hidden layers of width $2C$ and GELU activations, where $C$ is the frozen feature dimension ($C=384$ for HiViT-S). The predicted feature is L2-normalized and optimized using cosine distance to the target representation. All probes are trained for 50 epochs with AdamW, a learning rate of $10^{-3}$, weight decay of $10^{-4}$, and batch size 64. The checkpoint with the lowest validation loss is used for final evaluation.

\begin{figure*}[htb]
	\centering
	\includegraphics[width=\linewidth]{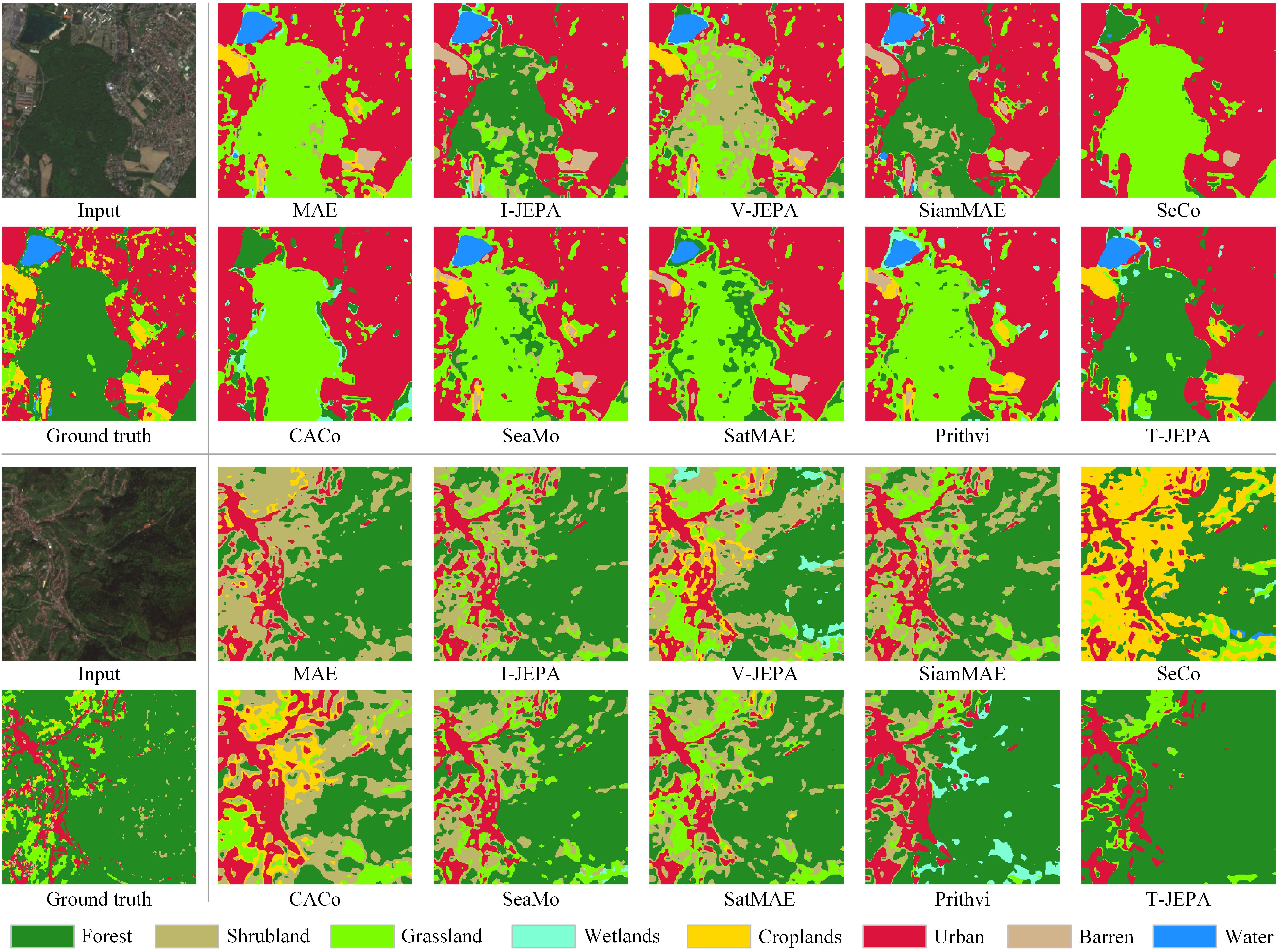}
	\caption{Qualitative comparison on DFC20 segmentation.}
	\label{fig:dfc}
\end{figure*}

We report three complementary metrics. PredSim is the cosine similarity between the predicted and target representations. Gain is the improvement over the identity baseline, defined as the difference between PredSim and the cosine similarity between the source and target representations, thereby reducing the advantage of temporally invariant features. Target R$@$1 measures whether the predicted representation retrieves the correct target timestamp among all five representations from the same location, with a random accuracy of 20\%. Finally, to evaluate sensitivity to temporal conditioning, we keep the trained predictor and source-target pairs fixed but randomly shuffle $\Delta t$ within the test batches and report the resulting retrieval accuracy.

\paragraph{Visualization of predicted latent.}

For the visualization analysis in Figure~4 of the main paper, we separately train a visualization-oriented variant of T-JEPA. Specifically, we remove the LayerNorm applied to future latent targets~\cite{ijepa} and the per-patch standardization applied to pixel reconstruction targets~\cite{mae}. These operations follow the default implementations of I-JEPA and MAE, respectively, and were omitted from the main Method section for brevity. The former depends on statistics of the ground-truth target latents, while the latter requires pixel statistics of the ground-truth input image. When the target observation is unavailable, these statistics cannot be obtained, preventing the predicted latent states from being directly decoded into pixel space. Removing both normalization operations during training enables the reconstruction decoder to directly map predicted latent states to images. In addition, the reconstruction branch is supervised on all patches rather than only the masked patches to improve visualization quality. As discussed in the original I-JEPA and MAE works, these normalization operations mainly stabilize optimization and do not constitute the core learning mechanism or introduce additional target information.

\begin{figure*}[htb]
	\centering
	\includegraphics[width=\linewidth]{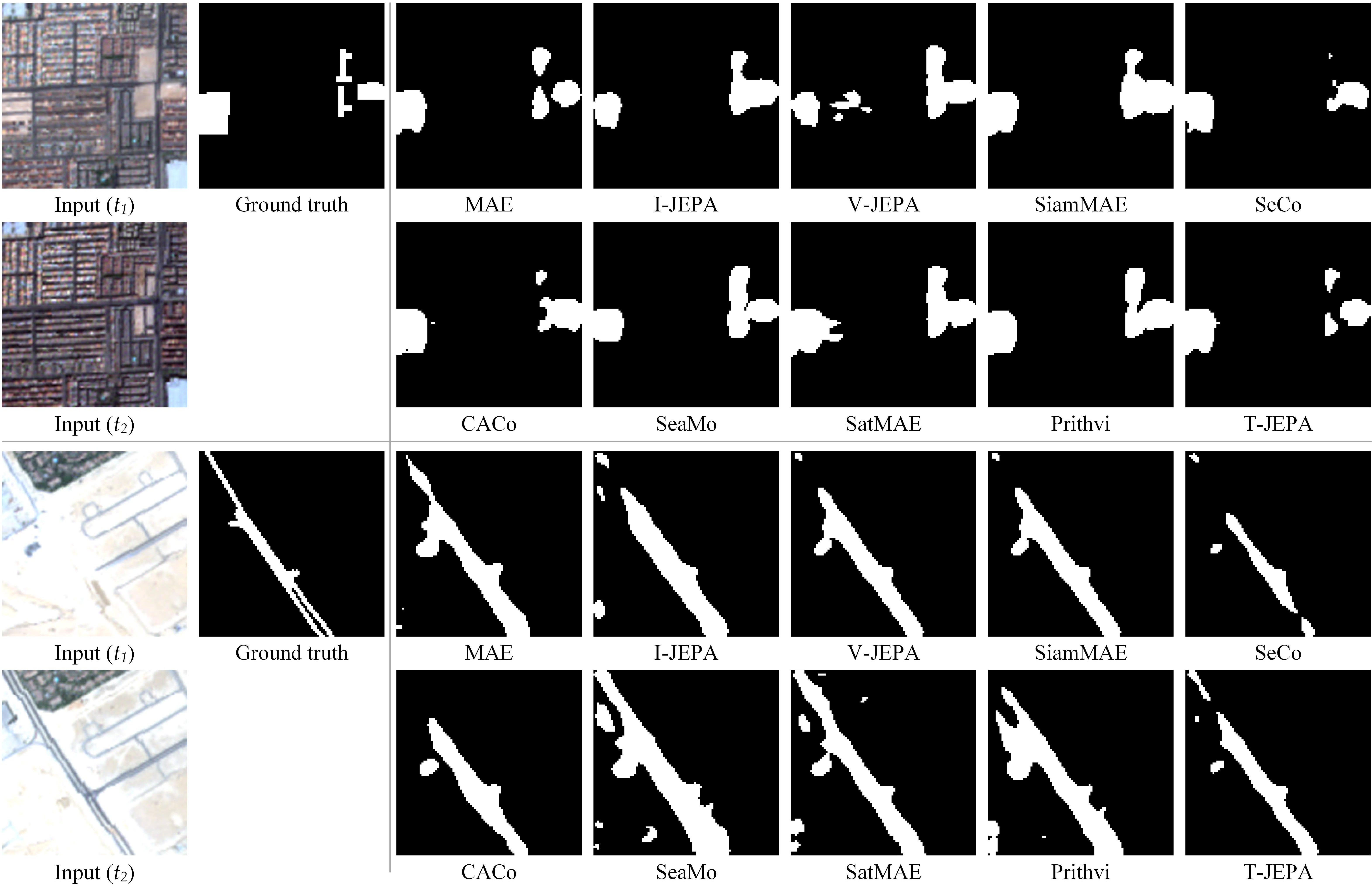}
	\caption{Qualitative comparison on OSCD change detection.}
	\label{fig:oscd}
\end{figure*}

After training, the model can predict latent states at a series of future intervals from a single ground-truth anchor observation and decode them into pixel space. Specifically, we randomly mask $50\%$ of the anchor patches and specify twelve approximately monthly target intervals, i.e., roughly 30, 60, $\ldots$, and 330 days. When a ground-truth observation is available within a predicted month, its actual acquisition date replaces the approximate monthly date to provide a more accurate time gap. The prediction process uses only the anchor observation and the specified target intervals; no subsequent images are provided to the model.

\subsection{Additional Results}

\paragraph{Qualitative results for downstream tasks.}

Figures~\ref{fig:pastis}, \ref{fig:dfc}, and \ref{fig:oscd} present qualitative comparisons on semantic segmentation and change detection. These visualizations complement the quantitative results by examining spatial coherence, boundary delineation, and the localization of small or fragmented regions. Across the displayed examples, T-JEPA generally produces predictions that more closely follow the ground-truth structures, with clearer boundaries and fewer spurious responses than the compared methods.

\paragraph{Ablation of the temporal prediction loss weight.}

\begin{table}[tbp]
    \centering
    \begin{tabular}{c|ccc}
    $\lambda$ & BEN & PASTIS & DFC20 \\ 
    \Xhline{1pt}
    0. & 60.5 & 46.8 & 38.3 \\
    0.01 & 62.6 & 48.0 &  38.0 \\
    0.1 & 64.9 & 50.1 & 37.7 \\ 
    0.5 & \textbf{66.0} & \textbf{50.7} & \textbf{39.4} \\
    1.0 & 65.5 & 50.3 & \textbf{39.4} \\
    5.0 & 65.4 & 49.5 & 38.2 \\
    
    \end{tabular}
    \caption{Ablation study on prediction loss weight.}
    \label{tab:lambda}
\end{table}

Table~\ref{tab:lambda} studies the trade-off controlled by the prediction weight $\lambda$. Weak prediction supervision provides limited temporal regularization, whereas excessively large weights may over-constrain the representation toward predictable temporal components at the expense of complementary visual information. Performance peaks at $\lambda=0.5$ and decreases slightly with larger values, highlighting the need to balance temporal predictability with spatial representation learning.

\begin{table*}[htbp]
\centering
\begin{tabular}{lrrrrrr}
    \toprule
    Method
    & Params & FLOPs/seq. & GPU time & Throughput & Peak memory & Wall-clock \\
    & (M) & (G) & (ms/iter) & (seq./s) & (GB) & (s/epoch) \\
    \midrule
    MAE & 49.04 & 45.24 & 142.24 & 224.98 & 9.57 & 50.15 \\
    T-JEPA & 64.06 & 145.16 & 267.52 & 119.62 & 14.14 & 60.18 \\
    \midrule
    Relative change & $+30.64\%$ & $+220.90\%$ & $+88.08\%$ & $-46.83\%$ & $+47.83\%$ & $+20.00\%$ \\
    \bottomrule
\end{tabular}
\caption{Pre-training cost of T-JEPA compared with MAE. FLOPs are reported per four-observation sequence. GPU time and throughput exclude data loading, whereas wall-clock time covers the complete training pipeline.}
\label{tab:eff_benchmark}
\end{table*}

\begin{table*}[htbp]
\centering
\begin{tabular}{lrrr}
    \toprule
    Component & Added FLOPs/seq. (G) & Share of added FLOPs & Forward CUDA time (ms/iter) \\ \midrule
    EMA target-encoder forward  & 72.35 & $72.40\%$ & 49.80 \\
    Prediction Transformer blocks & 27.14 & $27.16\%$ & 18.16 \\
    Prediction head & 0.35 & $0.35\%$ & 0.60 \\
    Prediction preparation & 0.09 & $0.09\%$ & 1.03 \\
    Time-gap embedder & $<0.01$ & $<0.01\%$ & 0.24 \\
    EMA parameter update & --  & -- & 9.25 \\
    \midrule
    Total additional forward FLOPs & 99.93 & $100.00\%$ & -- \\
    \bottomrule
\end{tabular}
\caption{Component-wise breakdown of the additional pre-training computation introduced by T-JEPA. CUDA times measure the corresponding forward operations, except for the separately reported EMA parameter update.}
\label{tab:eff_breakdown}
\end{table*}

\paragraph{Computational overhead analysis.}

We benchmark T-JEPA and its MAE counterpart under identical settings. Both models are trained on a single GPU using 2,000 samples, a batch size of 32 four-observation sequences, mixed precision, and FlashAttention, without activation checkpointing. The first epoch serves as warm-up, and measurements are collected during the second epoch. GPU time excludes data loading and augmentation, whereas wall-clock time covers the complete training pipeline.

As shown in Table~\ref{tab:eff_benchmark}, T-JEPA increases the number of trainable pre-training parameters by 30.64\% and forward computation by 99.93 GFLOPs per sequence, corresponding to a 220.90\% increase over MAE. Table~\ref{tab:eff_breakdown} shows that this overhead is dominated by the EMA target-encoder forward and prediction Transformer, which account for 72.40\% and 27.16\% of the added FLOPs, respectively. Together, they contribute approximately 99.6\% of the additional computation. By contrast, prediction preparation, the prediction head, and time-gap conditioning collectively account for less than 0.5\%. In particular, the time-gap embedder adds less than 0.01 GFLOPs and 0.24 ms per iteration.

The GPU training time increases by 88.08\%, from 142.24 to 267.52 ms per iteration. Component profiling attributes 49.80 ms to the EMA target forward and 20.03 ms to the prediction operations, including 18.16 ms for the Transformer blocks. The remaining overhead includes backpropagation through the prediction branch, optimization of its additional parameters, and the 9.25-ms EMA update. When the full data pipeline is included, epoch wall-clock time increases by only 20.00\%, from 50.15 to 60.18 seconds. Moreover, the EMA encoder and prediction branch are discarded after pre-training, so T-JEPA introduces no additional parameters or computation relative to MAE during downstream inference.
            
\end{document}